\documentclass[letterpaper]{article}

\usepackage[preprint]{aaai2027}
\usepackage[hyphens]{url}
\usepackage{graphicx}
\usepackage{natbib}
\usepackage{caption}
\usepackage{booktabs}
\usepackage{tabularx}
\usepackage{array}
\usepackage{amsmath,amssymb,amsfonts,amsthm}

\title{DualCert: A Solver for the Traveling Salesman Problem with Constraint-Coupled Learning}

\newcommand{\ArxivAuthorName}{Yancheng Song\textsuperscript{\rm 1}, Yongzhi Qi\textsuperscript{\rm 2}, Wei Qi\corresponding\textsuperscript{\rm 3}, Zuo-Jun Max Shen\textsuperscript{\rm 4}}
\newcommand{\ArxivAuthorAffiliation}{%
\textsuperscript{\rm 1} Department of Industrial Engineering, Tsinghua University, Beijing 100084, China,
songyc25@mails.tsinghua.edu.cn,\\
\textsuperscript{\rm 2} JD.com, Beijing, China, 101111, qiyongzhil@jd.com,\\
\textsuperscript{\rm 3} Department of Industrial Engineering, Tsinghua University, Beijing 100084, China,
qiw@tsinghua.edu.cn,\\
\textsuperscript{\rm 4} College of Engineering, UC Berkeley, Berkeley, CA 94720, USA; Faculty of Engineering,
Faculty of Business and Economics, University of Hong Kong, Hong Kong, China,
maxshen@berkeley.edu}

\author{\ArxivAuthorName}
\affiliations{\ArxivAuthorAffiliation}

\theoremstyle{plain}
\newtheorem{theorem}{Theorem}
\newtheorem{proposition}{Proposition}

\newcommand{\R}{\mathbb{R}}
\newcommand{\OPT}{\operatorname{OPT}}
\newcommand{\Diag}{\operatorname{Diag}}

\begin{document}
\maketitle

\begin{abstract}
Large traveling salesman problem (TSP) instances require a solver to allocate limited computation without weakening the validity of its reported outputs.
Existing neural--operations-research (OR) hybrids usually predict guidance, such as edge scores, penalties, or warm starts, without requiring each learned transition to satisfy constraints discovered during search.
DualCert introduces \emph{constraint-coupled learning}, under which current degree equations and dynamically separated subtour-elimination constraints (SECs) define the geometry of each learned transition.
At each refinement, the degree equations and selected, strictly satisfied SEC equations, represented with positive slacks, define an iterate-dependent primal-slack Karush--Kuhn--Tucker (KKT) manifold.
Repaired dual variables provide the reduced-cost component of a local cost field, and violated SEC rows provide its penalty component.
An exact constrained mirror-descent step maps each finite input state to a positive output on the same primal-slack KKT manifold.
On each region where the selected rows and deterministic ties remain fixed, implicit differentiation maps parameter perturbations into the manifold tangent space, while the local-cost-field derivative reuses the forward constraint operator.
The terminal edge state determines learned allocation across Held--Karp ascent, candidate-graph edge tests, and tour construction under a fixed computation budget.
Deterministic verification uses original-cost recomputation to accept only verified candidate-graph lower bounds and verified edge decisions.
On 1,000 held-out TSP1000 instances, DualCert attains a mean tour-cost gap of \(0.0573\%\) from Lin--Kernighan--Helsgaun version 3 (LKH-3) reference tours in \(9.55\) batch-amortized seconds per instance. It returns a verified candidate-graph lower bound for every instance and achieves \(81.46\%\) edge-decision coverage.
The mean gap is \(67.1\%\) smaller than the reported NeuroLKH mean gap.
Overall, optimization constraints define the edge state and learning geometry, while deterministic verification preserves the stated validity of reported lower bounds and edge decisions.
\end{abstract}

\section{Introduction}

The traveling salesman problem (TSP) asks for a minimum-cost Hamiltonian cycle, namely a closed route that visits every city exactly once.
A feasible tour upper-bounds the optimum, a lower bound excludes cheaper admissible tours, and edge tests identify edges that cannot or must appear in tours below a cost threshold.
Large instances permit only a subset of these calculations, making the strongest combined output within a finite computation budget the practical target.

Classical operations research (OR) provides the valid calculations required for these outputs.
Branch-and-cut enforces degree equations, which require one incoming and one outgoing tour edge per city, and separates subtour-elimination constraints (SECs), which exclude disconnected cycles~\citep{held1970traveling,applegate2006traveling}.
Held--Karp ascent adjusts node potentials, which are per-city cost corrections, and computes one-trees as lower bounds. Each one-tree consists of a spanning tree on the nonroot cities plus two edges incident to the root.
Lin--Kernighan--Helsgaun version 3 (LKH-3) improves tours through variable-length exchanges~\citep{helsgaun2000effective}. These valid routines compete for the same finite budget, so the solver must decide which calculations to execute.

Neural TSP methods improve finite-budget search by constructing tours, estimating promising edges, predicting penalties, selecting candidate sets, or choosing solver actions~\citep{vinyals2015pointer,joshi2019efficient,sun2023difusco,ma2021dact}.
Neural-OR hybrids learn solver scores or warm starts~\citep{xin2021neurolkh,xin2021vsrlkh,ye2024glop,gevers2024learning}, while differentiable layers embed fixed continuous programs~\citep{amos2017optnet,agrawal2019differentiable,donti2021dc3,liang2024homeomorphic}.
During large-instance search, however, new SEC rows appear as the state evolves, and an unconstrained neural proposal need not satisfy the rows currently in force.
Repairing only the final state therefore cannot make earlier transitions or their training derivatives obey those rows.

To address these limitations, DualCert follows one design principle: OR constraints govern learned transitions and derivatives, learning allocates computation, and deterministic routines determine reportable outputs.
We call this principle \emph{constraint-coupled learning}.
DualCert realizes the principle through four connected mechanisms. A sparse \emph{candidate graph} supports edge preferences, Held--Karp lower bounds, and edge tests, while the \emph{complete graph} of all city pairs supports tour completion and local search.

The computational sequence begins with a constraint-defined state.
The \emph{edge state} assigns positive preferences to \emph{candidate arcs}, directed copies of candidate edges, rather than a discrete tour.
Degree equations and selected, strictly satisfied SECs with positive slacks define the \emph{primal-slack Karush--Kuhn--Tucker (KKT) manifold}. This manifold contains positive states satisfying the selected equations; it is not the set of complete KKT optima.
\emph{Repaired dual variables} satisfy sign and stationarity conditions. Their reduced costs combine with penalties from violated SEC rows to define the \emph{local cost field}.

This geometry creates a finite-step requirement.
Tangent motion preserves the equations only to first order, so a finite update can leave the manifold.
The \emph{exact constrained mirror-descent step} minimizes the inner product between the local cost field and the updated state, plus a negative-entropy Bregman divergence---a relative-change measure for positive states---subject to the current equations.
Exact arithmetic preserves positivity and these equations, while numerical solution meets a prescribed tolerance.

Training differentiates the same finite transition used in the forward pass.
Within a region where the selected SEC rows and deterministic ties remain fixed, implicit differentiation maps parameter changes into the equation-preserving tangent space. The local-cost-field derivative uses the same constraint operator as the forward solve.

\begin{figure*}[t]
\centering
\includegraphics[width=0.95\textwidth]{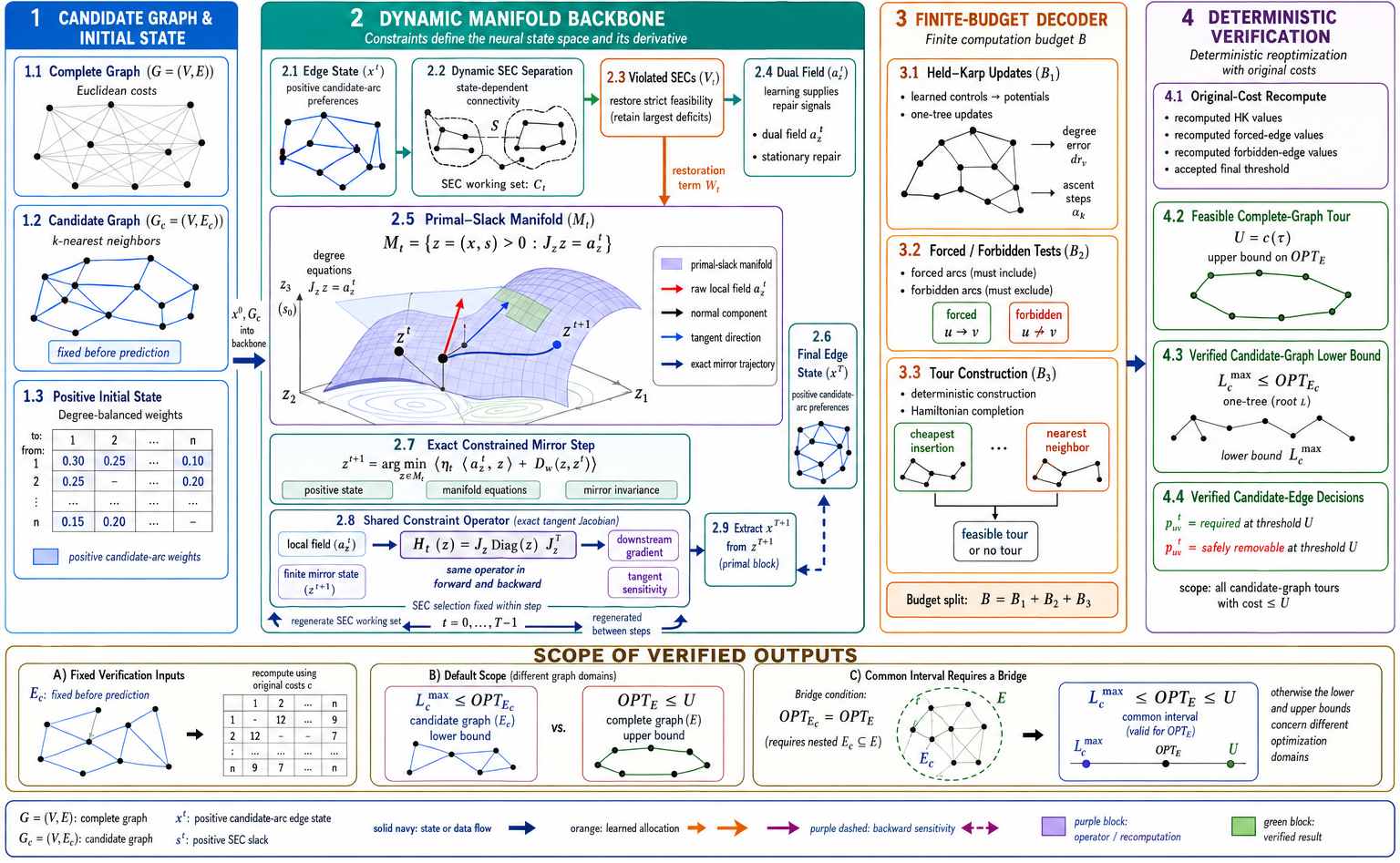}

\caption{DualCert couples learned computation to dynamic TSP constraint geometry. Degree and manifold SEC equations define the primal-slack KKT manifold, while repaired dual variables and violated SEC rows define the local cost field. The exact constrained mirror-descent step keeps the fixed-selection parameter derivative tangent to the manifold, while its local-cost-field derivative reuses the forward constraint operator. The terminal state allocates finite-budget computation, and deterministic verification determines which candidate-graph lower bounds and edge decisions are reportable.
}
\label{fig:framework}
\end{figure*}

The terminal state then controls a \emph{finite-budget decoder}.
\emph{Learned allocation} predicts initial Held--Karp node potentials and ascent steps and orders edge tests and construction.
Construction followed by a fixed LKH-3 configuration returns a complete-graph tour.
\emph{Deterministic verification} recomputes each record from the original costs and accepts an edge decision only when its restricted one-tree bound exceeds the returned tour-cost threshold.
Forced and forbidden tests then mark edges safely removable and required, respectively, for candidate-graph tours of cost at most that threshold.
Figure~\ref{fig:framework} summarizes this sequence.

On 1,000 held-out TSP1000 instances, DualCert returns tours in \(9.55\) batch-amortized seconds per instance with a mean gap of \(0.0573\%\) from the LKH-3 reference tours. Here the gap is the percentage difference between the returned and reference tour costs. This mean gap is \(67.1\%\) smaller than the NeuroLKH value measured on the same instances. DualCert also returns a verified candidate-graph lower bound for every instance and resolves \(81.46\%\) of candidate edges at the returned-tour threshold.

In summary, DualCert makes four contributions:
\begin{enumerate}
    \item \textbf{Constraint-defined state geometry.}
    Current degree and SEC equations govern neural refinement through the primal-slack KKT manifold, while repaired dual variables and violated SEC rows define the local cost field.

    \item \textbf{Exact finite-step and backward coupling.}
    The exact constrained mirror-descent step preserves manifold feasibility at finite step size. On fixed-selection regions, its parameter derivative remains tangent to the manifold, while its local-cost-field derivative reuses the forward constraint operator.

    \item \textbf{Validity-preserving finite-budget allocation.}
    Learned allocation controls Held--Karp ascent, edge testing, and construction. Regardless of the allocation, fixed verification conditions determine which candidate-graph lower bounds and edge decisions are reportable.

    \item \textbf{Finite-budget evidence.}
    DualCert achieves a \(0.0573\%\) TSP1000 mean gap, returns a verified candidate-graph lower bound for every instance, and reaches \(81.46\%\) edge-decision coverage.
\end{enumerate}

\section{Related Work}
\label{sec:related}

\paragraph{Classical TSP Optimization}
Classical TSP optimization supplies computations with explicit validity conditions.
Branch-and-cut separates SECs, Held--Karp ascent produces candidate-graph lower bounds, and restricted one-trees support candidate-graph edge tests~\citep{held1970traveling,applegate2006traveling}.
LKH-3 improves complete-graph tours through variable-length exchanges~\citep{helsgaun2000effective}.
Large instances admit more lower-bound calculations, candidate-graph edge tests, and local moves than a finite budget can execute.
Therefore, DualCert uses learned allocation to select computations, while deterministic verification retains the classical validity conditions.
This need to prioritize valid calculations motivates neural TSP methods.

\paragraph{Neural TSP Methods}
Neural TSP methods use constructive policies, edge heatmaps, diffusion models, or improvement operators to focus finite computation~\citep{vinyals2015pointer,joshi2019efficient,sun2023difusco,ma2021dact}.
Hybrid methods retain a classical solver and learn solver inputs.
NeuroLKH predicts edge scores and node penalties, VSR-LKH learns search choices, GLOP learns decomposition, and learned multipliers initialize Held--Karp ascent~\citep{xin2021neurolkh,xin2021vsrlkh,ye2024glop,gevers2024learning}.
These hybrid interfaces show that learned inputs can improve a classical solver.
In most such interfaces, however, current degree and SEC equations do not define either the learned proposal's finite internal transition or the derivative of that transition.
DualCert instead makes the current equations define both the finite edge-state transition and fixed-selection parameter sensitivity.

\paragraph{Differentiable Constrained Optimization}
Differentiable constrained optimization provides the closest comparison because it explicitly couples learning to solver constraints.
Prior methods differentiate fixed programs or map proposals to feasible outputs~\citep{amos2017optnet,agrawal2019differentiable,mena2018gumbel,wang2019satnet,ferber2020mipaal,vlastelica2020differentiation,donti2021dc3,liang2024homeomorphic}.
DOGE-Train preserves dual feasibility and lower-bound monotonicity in differentiable decomposition~\citep{abbas2024dogetrain}.
DualCert instead addresses a finite refinement trajectory whose SEC working set changes with the iterate.
Its exact constrained mirror-descent step closes the finite-update gap left by tangent motion. The resulting terminal state then drives learned allocation, after which deterministic verification determines reportable outputs.

\section[Problem Scope and Solver Outputs]{Problem Scope and\\Solver Outputs}
\label{sec:scope}

As summarized in Figure~\ref{fig:framework}, DualCert uses the candidate graph for repeated calculations and the complete graph for tour completion, so every reported output requires an explicit graph scope.

\paragraph{The complete graph and candidate graph have distinct roles.}
Let \(G=(V,E)\) be the complete graph on \(n=|V|\geq3\) Euclidean cities, and let
\(c=(c_e)_{e\in E}\in\R^{|E|}\) be the original cost vector, with \(c_e\geq0\)
for every undirected edge \(e\in E\).

Repeated edge-state, one-tree, and edge-test calculations on the complete edge
set \(E\) are expensive. DualCert therefore fixes a sparse candidate graph
\(G_c=(V,E_c)\) and performs these calculations on \(E_c\). This restriction
reduces memory use and one-tree computation, while the complete graph remains
available for final tour completion and local search. The associated directed
candidate-arc set is
\[
\vec E_c=\{(i,j),(j,i):\{i,j\}\in E_c\},
\qquad
m=|\vec E_c|.
\]
Each candidate edge \(\{i,j\}\in E_c\) has two directed copies,
\((i,j),(j,i)\in\vec E_c\). A cycle cover
\(\Gamma\subseteq\vec E_c\) gives every city exactly one incoming and exactly
one outgoing candidate arc.

Preprocessing removes arcs that belong to no cycle cover, inserts a deterministic cycle cover when necessary, and requires a candidate-graph one-tree for a fixed Held--Karp root \(v_\star\in V\).
The cycle-cover condition supports positive degree balance, while the one-tree condition supports every candidate-graph Held--Karp calculation.
Neither condition asserts that \(G_c\) contains a complete-graph optimal tour.
Preprocessing then freezes \(G_c\) and the total tie order \(\prec_{\rm tie}\) for equal-valued choices.

\paragraph{The two graph scopes induce two optimization targets.}
The complete-graph and candidate-graph optima are
\begin{equation}
\begin{aligned}
\OPT_E
&=\min_{\substack{\tau\text{ is a Hamiltonian cycle}\\\tau\subseteq E}}
\sum_{e\in\tau}c_e,\\
\OPT_{E_c}
&=\min_{\substack{\tau\text{ is a Hamiltonian cycle}\\\tau\subseteq E_c}}
\sum_{e\in\tau}c_e.
\end{aligned}
\label{eq:scope-opt}
\end{equation}
Here and below, \(\min\varnothing:=+\infty\).
Every candidate-graph tour is also a complete-graph tour, which gives \(\OPT_E\leq\OPT_{E_c}\).
The \emph{scope condition} is \(\OPT_E=\OPT_{E_c}\), and this equality holds exactly when \(G_c\) contains a complete-graph optimal tour.

\paragraph{The edge state records positive preferences within the candidate-graph scope.}
The \(T=6\) refinements use \(t=0,\ldots,T-1\), and each state matrix \(X^t\), for \(t=0,\ldots,T\), stores one positive preference \(X^t_{ij}\) per candidate arc.
The two directed copies of each candidate edge permit orientation-specific preferences and separate incoming and outgoing degree equations, while both copies still represent the same undirected edge.
Vectorization gives the edge state
\(
x^t=\operatorname{vec}(X^t)\in\R_{++}^{m},
\)
where \(\R_{++}^{m}\) is the set of positive \(m\)-vectors.
The continuous state \(x^t\) encodes edge preferences rather than a discrete tour, and \(x^T\) enters the finite-budget decoder.

\paragraph{The fixed budget determines the reportable outputs.}
Let \(\mathcal B>0\) denote the fixed computation budget, and define the original cost of any tour \(\tau\) by \(c(\tau)=\sum_{e\in\tau}c_e\).
Within \(\mathcal B\), DualCert returns a complete-graph tour \(\tau\) and sets the reporting threshold to \(U=c(\tau)\). It reports safely removable candidate edges \(F^{\mathrm{out}}\subseteq E_c\) and required candidate edges \(F^{\mathrm{in}}\subseteq E_c\). When at least one unrestricted lower-bound record passes deterministic verification, it additionally reports the verified candidate-graph lower bound \(L_c^{\max}\).
The graph scopes imply \(\OPT_E\leq U\). Whenever \(L_c^{\max}\) is reported, they also imply \(L_c^{\max}\leq\OPT_{E_c}\).
Section~\ref{sec:decoder} defines the original-cost recomputation that produces \(L_c^{\max}\), \(F^{\mathrm{out}}\), and \(F^{\mathrm{in}}\).

\section{Constraint-Coupled Backbone}
\label{sec:backbone}

The backbone learns a short edge-state trajectory whose admissible finite transitions are defined by the current degree and SEC equations.
\paragraph{The candidate-support interior supplies the initial edge state.}
Let \(A\in\R^{r_A\times m}\) be a fixed full-row-rank basis of the incoming and outgoing degree equations, where \(r_A\) is the number of independent equations, and let \(b\in\R^{r_A}\) be the corresponding right-hand side.
The preprocessing in Section~\ref{sec:scope} ensures that the candidate-support interior
\(
\mathcal D_c=\{x\in\R_{++}^{m}:A x=b\}
\)
is nonempty.

Indeed, for each retained arc, choose a cycle cover that contains it and average the resulting covers. The averaged vector gives positive mass to every retained arc and satisfies the incoming and outgoing degree equations.
Define the positive distance initialization \(w^0\in\R_{++}^{m}\) by \(w^0_{ij}=(1+c_{\{i,j\}})^{-1}\) for \((i,j)\in\vec E_c\).
For a positive vector \(z\), let \(\log z\) act componentwise, let \(\mathbf 1\) be the matching all-ones vector, and let \(\langle\cdot,\cdot\rangle\) denote the Euclidean inner product.
Negative entropy \(\omega\) is
\(
\omega(z)=\langle z,\log z-\mathbf 1\rangle.
\)
For positive vectors \(u\) and \(z\) of equal dimension, the associated Bregman divergence \(D_\omega\) is
\[
\begin{aligned}
D_\omega(u,z)
=\omega(u)-\omega(z)-\langle\nabla\omega(z),u-z\rangle.
\end{aligned}
\]
The entropy projection
\(
x^0=\arg\min_{x\in\mathcal D_c}D_\omega(x,w^0)
\)
therefore initializes the trajectory with \(x^0\in\mathcal D_c\).

\paragraph{Deterministic separation forms the SEC working set.}
Degree balance alone permits disconnected cycle covers, so the backbone adds connectivity equations exposed by the evolving state. At refinement \(t\), deterministic separation forms the \emph{SEC working set} \(\mathcal C_t\). Later refinements regenerate this set because a new state \(x^t\) can expose a previously unrepresented connectivity violation.
Each index \(q\in\mathcal C_t\) identifies a node subset. The row vector \(B_q\in\R^{1\times m}\) sums the candidate-arc mass leaving that subset, and \(h_q=1\). Thus, \(B_qx\geq h_q\) requires at least one unit of outgoing mass. The fixed order \(\prec_{\rm tie}\) resolves separation ties.
The \emph{manifold SEC set} \(\mathcal S_t\) contains strictly satisfied rows, whereas the \emph{violated subset} \(\mathcal V_t\) contains violated rows:
\[
\begin{aligned}
\mathcal S_t&=\{q\in\mathcal C_t\colon B_q x^t>h_q\},\\
\mathcal V_t&=\{q\in\mathcal C_t\colon B_q x^t<h_q\}.
\end{aligned}
\]
A boundary row with \(B_q x^t=h_q\) belongs to neither set. Consequently, \(\mathcal S_t\) contains exactly the rows with positive manifold slacks, and \(\mathcal V_t\) supplies exactly the rows used by the local penalty. For \(\mathcal Q\subseteq\mathcal C_t\), \(B_{\mathcal Q}\) and \(h_{\mathcal Q}\) stack the corresponding rows and right-hand sides.

\paragraph{The manifold SEC set defines the primal-slack KKT manifold.}
Let \(I_{|\mathcal S_t|}\) be the identity matrix of order \(|\mathcal S_t|\), and let the zero block below have \(r_A\) rows and \(|\mathcal S_t|\) columns.
The positive slack vector \(s^t\), augmented primal-slack state \(z^t\), manifold equation matrix \(J_t\), and manifold right-hand side \(q_t\) are
\begin{equation}
\begin{aligned}
s^t&=B_{\mathcal S_t}x^t-h_{\mathcal S_t}>0,
&z^t&=\begin{bmatrix}x^t\\s^t\end{bmatrix},\\
J_t&=\begin{bmatrix}A&0\\B_{\mathcal S_t}&-I_{|\mathcal S_t|}\end{bmatrix},
&q_t&=\begin{bmatrix}b\\h_{\mathcal S_t}\end{bmatrix}.
\end{aligned}
\label{eq:primal-slack-state}
\end{equation}
These objects define the \emph{primal-slack KKT manifold}
\begin{equation}
\mathcal M_t=
\left\{z\in\R_{++}^{m+|\mathcal S_t|}:J_t z=q_t\right\}.
\label{eq:manifold}
\end{equation}

The slack coordinates turn selected inequalities into equalities with positive margins, so finite updates on \(\mathcal M_t\) cannot cross the corresponding SEC boundaries during refinement \(t\).
The KKT qualifier refers to the pairing of these primal equations with the repaired dual sign and stationarity equations introduced below. The manifold itself contains only the primal state and positive slack coordinates; it does not impose complementary slackness or characterize a complete optimum.
Full row rank of \(A\) and one private slack coordinate per manifold SEC row make \(J_t\) full row rank. Here \(\ker J_t\) denotes the kernel of \(J_t\).
Since \(J_tz=q_t\) defines primal-slack feasibility, its tangent space at every \(z\in\mathcal M_t\) is
\[
T_z\mathcal M_t
=
\ker J_t
=
\left\{
\xi\in\R^{m+|\mathcal S_t|}
:
J_t\xi=0
\right\}.
\]

\paragraph{Repaired dual variables define the reduced-cost term.}
The manifold specifies admissible states, while repaired dual variables define the local reduced-cost term that guides the update.
The directed candidate-arc cost vector \(\bar c\in\R^m\) copies \(c_e\) to both arcs of each \(e\in E_c\), and \(\phi\) denotes all trainable parameters.
At refinement \(t\), sign and stationarity repair produces four objects: the degree multiplier \(\lambda_\phi^t\in\R^{r_A}\), the SEC multiplier \(\mu_\phi^t\in\R^{|\mathcal C_t|}\), the local-upper-bound multiplier \(\nu_\phi^t\in\R^m\), and the positive repaired reduced-cost term \(r_\phi^t\in\R^m\).
For the local upper bound \(x\leq\mathbf 1\), the repaired dual variables \((\lambda_\phi^t,\mu_\phi^t,\nu_\phi^t,r_\phi^t)\) satisfy
\begin{equation}
\begin{aligned}
\bar c-A&^\top\lambda_\phi^t-B_{\mathcal C_t}^\top\mu_\phi^t+\nu_\phi^t-r_\phi^t=0,\\
\mu_\phi^t&\geq0,
\qquad
\nu_\phi^t\geq0,
\qquad
r_\phi^t>0.
\end{aligned}
\label{eq:repair}
\end{equation}
Equation~\eqref{eq:repair} is the required output interface of the repair module. The analysis below uses these sign and stationarity conditions and does not require a particular internal parameterization of that module.
\paragraph{The violated subset completes the local cost field.}
The violated subset augments the repaired reduced costs with a differentiable SEC penalty.
The weight \(\beta_{\rm sec}\geq0\) scales this penalty.
For any vector \(u\), \([u]_+\) denotes the componentwise maximum of \(u\) and zero, and \(\|\cdot\|_2\) denotes the Euclidean norm.
For the following local derivative, hold the repaired dual variables and selected row sets fixed. Using \(0_{|\mathcal S_t|}\) for the zero vector matching the slack block, define the local objective \(\psi_t\), its edge-state gradient \(g_\phi^t\), and the augmented local cost field \(a_\phi^t\) by
\begin{equation}
\begin{aligned}
\psi_t(x)
&=(r_\phi^t)^\top x
+\frac{\beta_{\rm sec}}{2}
\left\|[h_{\mathcal V_t}-B_{\mathcal V_t}x]_+\right\|_2^2,\\
g_\phi^t
&=\nabla_x\psi_t(x^t),
a_\phi^t
=\begin{bmatrix}
g_\phi^t\\
0_{|\mathcal S_t|}
\end{bmatrix}.
\end{aligned}
\label{eq:local-field}
\end{equation}
Thus, repaired reduced costs and violated SEC rows determine the local cost field, while \(J_t\) determines admissible motion.

\paragraph{The constraint projector describes the infinitesimal admissible motion.}
The local cost field specifies an ambient change, while the manifold permits only changes that preserve the current equations to first order.
For a positive vector \(z\), let \(\Diag(z)\) be the diagonal matrix with diagonal \(z\).
At \(z^t\), define the state metric pair
\(
W_t=\Diag(z^t),
G_t=W_t^{-1}.
\)
For an augmented covector \(a\in\R^{m+|\mathcal S_t|}\), the \emph{constraint projector}
\begin{equation}
\mathcal P_t(a)=
W_t a-W_tJ_t^\top
(J_t W_t J_t^\top)^{-1}J_tW_t a
\label{eq:projector}
\end{equation}
is the unique \(G_t\)-metric projection of \(W_t a\) onto \(\ker J_t\).
Full row rank of \(J_t\) and positivity of \(W_t\) make \(J_t W_t J_t^\top\) positive definite.
Consequently, \(J_t \mathcal P_t(a)=0\) places \(\mathcal P_t(a)\) in \(T_{z^t}\mathcal M_t\).

\paragraph{Infinitesimal admissibility does not guarantee a finite admissible transition.}
Let \(\odot\), \(\oslash\), and \(\exp\) denote componentwise multiplication, division, and exponentiation, and let \(\eta_t>0\) be the refinement step size.
The raw multiplicative update
\(
\widetilde z_t(\eta_t)=
z^t\odot\exp\left(-\eta_t\mathcal P_t(a_\phi^t)\oslash z^t\right)
\)
does not generally preserve the equations at finite step size. Writing \(p_t=\mathcal P_t(a_\phi^t)\), its constraint residual has the expansion
\begin{equation}
\begin{aligned}
J_t \widetilde z_t(\eta_t)-q_t
&=\frac{\eta_t^2}{2}
J_t\left((p_t\odot p_t)\oslash z^t\right)\\
&\quad+O(\eta_t^3).
\end{aligned}
\label{eq:finite-drift}
\end{equation}
The term \(O(\eta_t^3)\) collects terms of order three and higher.
The first-order residual vanishes, but the finite residual generally remains nonzero.
Therefore, exact preservation requires a finite constrained map.

\paragraph{The exact constrained mirror-descent step preserves manifold feasibility at finite step size.}
Let \(z^{t+1,-}\) denote the output on \(\mathcal M_t\) before SEC regeneration, and let \(\rho_t\in\R^{r_A+|\mathcal S_t|}\) be the multiplier associated with the manifold equations.
The \emph{exact constrained mirror-descent step} is the solution of the constrained optimization problem below. Its stationarity equation gives the exponential form on the second line. For \(\rho\in\R^{r_A+|\mathcal S_t|}\), the third line defines the associated dual residual \(F_t(\rho)\), and the final line enforces feasibility:
\begin{equation}
\begin{aligned}
z^{t+1,-}
&=\arg\min_{z\in\mathcal M_t}
\left\{\eta_t\langle a_\phi^t,z\rangle+D_\omega(z,z^t)\right\},\\
z^{t+1,-}
&=z^t\odot
\exp\left(-\eta_ta_\phi^t-J_t^\top\rho_t\right),\\
F_t(\rho)
&=J_t\left[z^t\odot
\exp\left(-\eta_ta_\phi^t-J_t^\top\rho\right)\right]-q_t,\\
F_t(\rho_t)&=0.
\end{aligned}
\label{eq:mirror}
\end{equation}

The first line defines the exact constrained mirror-descent transition, while \(F_t(\rho_t)=0\) reduces its computation to a sparse dual solve. The next result establishes existence and uniqueness of this transition and nonsingularity of the associated dual system.

\begin{theorem}[Finite-step manifold invariance]
\label{thm:mirror}
For \(z^t\in\mathcal M_t\), finite \(a_\phi^t\), and \(0<\eta_t<\infty\), Equation~\eqref{eq:mirror} has a unique positive solution \(z^{t+1,-}\in\mathcal M_t\) and a unique multiplier \(\rho_t\).
Moreover, the Jacobian of \(F_t\) with respect to \(\rho\) satisfies
\[
-D_\rho F_t(\rho_t)
=
J_t\Diag(z^{t+1,-})J_t^\top.
\]
The displayed matrix is positive definite.
\end{theorem}

Thus, the update preserves the current manifold at finite step size, rather than only following a feasible tangent direction, and the positive-definite dual system makes the multiplier solve well posed.
The nonnegative closure of \(\mathcal M_t\) is compact because the degree equations bound \(x\) and the slack equations bound \(s\).
Strict convexity gives a unique minimizer on this closure, and the feasible direction toward \(z^t>0\) excludes every boundary minimizer.
Full row rank of \(J_t\) then gives multiplier uniqueness and the positive-definite dual system.
Differentiating Equation~\eqref{eq:mirror} at zero step size gives
\[
\left.
\frac{\partial z^{t+1,-}}{\partial\eta_t}
\right|_{\eta_t=0}
=
-\mathcal P_t(a_\phi^t),
\]
so the exact constrained mirror-descent step is the finite constrained map associated with the same infinitesimal geometry.

The exact constrained mirror-descent step also determines the backward sensitivity of the state update. Finite-step invariance alone does not guarantee that training differentiates the constrained transition used during inference.
A \emph{fixed-selection region} \(\Omega_t^{\rm sel}\) keeps \(\mathcal C_t\), \(\mathcal S_t\), \(\mathcal V_t\), and every other combinatorial choice unchanged, excluding each boundary \(B_q x=h_q\) for \(q\in\mathcal C_t\).
At the finite output, define
\[
z^+=z^{t+1,-},
\qquad
W^+=\Diag(z^+),
\]
and define the output constraint projector by
\begin{equation}
\mathcal P_t^+(a)=
W^+a-W^+J_t^\top
(J_t W^+ J_t^\top)^{-1}J_tW^+a.
\label{eq:output-projector}
\end{equation}
The operators \(D_a\) and \(D_\phi\) denote differentiation with respect to the local cost field \(a_\phi^t\) and the trainable parameters \(\phi\), respectively.
The next result shows that forward correction and backward sensitivity obey the same constraint geometry.

\begin{theorem}[Shared forward--backward constraint operator]
\label{thm:jacobian}
For every local-cost-field perturbation \(v\in\R^{m+|\mathcal S_t|}\), the finite-update derivative satisfies
\begin{equation}
D_a z^+[v]=-\eta_t\mathcal P_t^+(v),
\qquad
J_t D_a z^+[v]=0.
\label{eq:field-jacobian}
\end{equation}
For any scalar downstream loss \(\mathcal L\), the corresponding vector--Jacobian product is
\(
\nabla_a\mathcal L
=-\eta_t\mathcal P_t^+\big(\nabla_{z^+}\mathcal L\big).
\)
On every \(\Omega_t^{\rm sel}\), the full parameter-to-state Jacobian satisfies
\[
\operatorname{range}(D_\phi z^{t+1,-})
\subseteq
T_{z^{t+1,-}}\mathcal M_t
=\ker J_t.
\]
\end{theorem}
The forward and local-cost-field backward solves therefore reuse \(J_t W^+ J_t^\top\).
Because \(J_t\) and \(q_t\) stay fixed, differentiating \(J_t z^{t+1,-}=q_t\) places every parameter-induced state change in \(\ker J_t\).
SEC regeneration occurs after the finite update and is not differentiated within a fixed-selection region \(\Omega_t^{\rm sel}\).

\paragraph{SEC regeneration preserves manifold membership across refinements.}
Let \(x^{t+1}\) be the edge-state block of \(z^{t+1,-}\).
The finite update preserves equality and positive slack for every row in \(\mathcal S_t\). Separation then forms \(\mathcal C_{t+1}\) and defines
\[
\begin{aligned}
\mathcal S_{t+1}
&=\{q\in\mathcal C_{t+1}\colon B_q x^{t+1}>h_q\},\\
s^{t+1}
&=B_{\mathcal S_{t+1}}x^{t+1}-h_{\mathcal S_{t+1}}>0.
\end{aligned}
\]
The regenerated state \(z^{t+1}=(x^{t+1},s^{t+1})\) lies in \(\mathcal M_{t+1}\). Applying this regeneration after every refinement, including \(t=T-1\), and using \(x^0\in\mathcal D_c\) as the base case gives \(z^t\in\mathcal M_t\) for \(t=0,\ldots,T\).

\paragraph{The numerical implementation verifies the modeled finite transition.}
The preceding guarantees use exact arithmetic, so the floating-point implementation must verify the realized manifold residual.
Let \(\widehat z^{t+1,-}>0\) be the numerical output and let \(\varepsilon_{\rm mir}>0\) be the prescribed residual tolerance for the exact constrained mirror-descent step.
The implementation accepts the output only when
\[
\left\|J_t \widehat z^{t+1,-}-q_t\right\|_2\leq\varepsilon_{\rm mir}
\]
and independent checks confirm positivity and satisfaction of the redundant degree equations.
Equation~\eqref{eq:mirror} is therefore exact in the model and solved to the prescribed floating-point tolerance.

\paragraph{Relative smoothness gives a sufficient descent condition.}
To ensure that a constraint-preserving update also decreases the local objective, we next use relative smoothness to derive a sufficient step-size condition.
Let \(\mathcal K_t\subset\mathcal M_t\) be a bounded positive region containing the states under consideration at refinement \(t\). For two states in this region, \(s_x\) and \(s_y\) denote the slack blocks paired with \(x\) and \(y\), respectively.
Let \(L_t>0\) satisfy
\begin{equation}
\begin{aligned}
\psi_t(y)\leq{}&\psi_t(x)
+\langle\nabla\psi_t(x),y-x\rangle\\
&+L_tD_\omega\big((y,s_y),(x,s_x)\big)
\end{aligned}
\label{eq:relative-smoothness}
\end{equation}
for all \((x,s_x),(y,s_y)\in\mathcal K_t\).
For the learned raw step \(\widehat\eta_t\in\R\), define
\[
\eta_t=\frac{\sigma(\widehat\eta_t)}{L_t},
\qquad
\sigma(u)=\frac{1}{1+\exp(-u)}.
\]
With this choice, the step in Equation~\eqref{eq:mirror} strictly decreases \(\psi_t\) whenever \(z^{t+1,-}\neq z^t\).

The terminal state connects constraint-coupled refinement to learned allocation of finite-budget decoder computation.
For each candidate edge \(e=\{i,j\}\in E_c\), define the terminal edge score
\(
\chi_e=\max\{X^T_{ij},X^T_{ji}\}.
\)
The scores \(\chi_e\) order tests and construction, while \(x^T\) parameterizes the remaining decoder controls.
Training composes these fixed-selection derivatives through the \(T\) refinements. When evaluating the gradient, it holds candidate-graph one-trees and other discrete decoder outputs fixed, so the stated derivative applies to the current selection trace.

\section[Finite-Budget Decoder and Deterministic Verification]{Finite-Budget Decoder and\\Deterministic Verification}
\label{sec:decoder}

The terminal state \(x^T\) parameterizes the learned finite-budget allocation.
For the trace executed under budget \(\mathcal B\), let \(K\geq1\) be the number of Held--Karp ascent updates and \(N_{\rm test}\geq0\) the number of candidate-graph edge tests.
The allocation
\[
\mathcal A_\phi(x^T,\mathcal B)
=
\left(
K,
N_{\rm test},
\pi^0,
\{\alpha_k\}_{k=0}^{K-1},
\prec_{\rm test},
\prec_{\rm con}
\right)
\]
contains the two trace lengths, the initial Held--Karp potential \(\pi^0\in\R^n\), positive ascent steps \(\alpha_k>0\), test order \(\prec_{\rm test}\), and construction order \(\prec_{\rm con}\).
The scores \(\chi_e\) induce both orders, with \(\prec_{\rm tie}\) resolving equal scores.
The decoder executes the \(K\) ascent updates and the first \(N_{\rm test}\) edge tests in \(\prec_{\rm test}\), so these outputs determine the finite computation trace. Deterministic verification then applies fixed acceptance conditions to the resulting records.

\paragraph{Held--Karp ascent turns learned controls into lower-bound records.}

Held--Karp ascent converts the allocated initial potential and step sizes into lower-bound records through deterministic candidate-graph one-tree minimization.
For \(k=0,\ldots,K-1\), the vector \(\pi^k\in\R^n\) contains the node potentials before update \(k\), and \(\alpha_k\) is the corresponding ascent step size. A fixed ascent routine starts from \(\pi^0\) and uses these steps to produce \(\pi^1,\ldots,\pi^K\).
Let \(\mathfrak O_c\) be the set of candidate-graph one-trees rooted at \(v_\star\). For \(\pi\in\R^n\), define the value of \(\mathcal O\in\mathfrak O_c\) and the corresponding minimum by
\begin{equation}
\begin{aligned}
\ell_c(\pi,\mathcal O)
&=\sum_{\{i,j\}\in\mathcal O}
\big(c_{\{i,j\}}+\pi_i+\pi_j\big)
-2\sum_{i\in V}\pi_i,\\
L_c(\pi)&=\min_{\mathcal O\in\mathfrak O_c}
\ell_c(\pi,\mathcal O).
\end{aligned}
\label{eq:held-karp-value}
\end{equation}
For \(k=0,\ldots,K\), deterministic minimization at \(\pi^k\) returns an optimizer \(\mathcal O_k\) and the value \(L_c(\pi^k)\leq\OPT_{E_c}\).
The record \(H_k=(\pi^k,\mathcal O_k,L_c(\pi^k))\) stores the calculation, and \(\mathcal H=\{H_k\}_{k=0}^{K}\) is the unrestricted record set.
These records provide lower-bound candidates, and \(\prec_{\rm test}\) selects restricted calculations.

\paragraph{Restricted one-tree tests produce records for later verification.}
The record set \(\mathcal R\) contains outputs from executed forced- or forbidden-edge tests that return a restricted candidate-graph one-tree.
Each record \(R\in\mathcal R\) has the form
\[
R=(\pi_R,e_R,\delta_R,\mathcal O_R).
\]
The entries are the selected potential \(\pi_R\in\R^n\), tested edge \(e_R\in E_c\), restriction type \(\delta_R\in\{\mathrm{force},\mathrm{forbid}\}\), and returned restricted one-tree \(\mathcal O_R\subseteq E_c\).
A forced-edge test minimizes over \(\mathfrak O_c^{\mathrm{force}}(e_R)=\{\mathcal O\in\mathfrak O_c:e_R\in\mathcal O\}\), while a forbidden-edge test minimizes over \(\mathfrak O_c^{\mathrm{forbid}}(e_R)=\{\mathcal O\in\mathfrak O_c:e_R\notin\mathcal O\}\).
The stored objects support later original-cost recomputation.
The same scores then order tour construction.

\paragraph{Construction converts \(\prec_{\rm con}\) into a complete-graph tour.}
Following \(\prec_{\rm con}\), construction accepts a candidate edge only when endpoint degrees remain at most two and the edge joins distinct components.
The accepted set \(P\subseteq E_c\) is therefore a path forest, namely a collection of vertex-disjoint paths.
Complete-graph edges close \(P\) into a Hamiltonian cycle \(\tau_0\), and fixed LKH-3 local search improves \(\tau_0\) to the returned tour \(\tau\).
The returned tour's recomputed cost defines the common threshold \(U\).

\paragraph{Deterministic verification converts records into reportable outputs.}
Construction first sets the common tour-cost threshold \(U=c(\tau)\). For each \(H_k\in\mathcal H\), the verifier rebuilds Equation~\eqref{eq:held-karp-value} from the original costs and reruns the deterministic minimum-one-tree calculation.
Let \(\mathcal H_{\rm ver}\subseteq\mathcal H\) contain the records whose stored one-tree and lower-bound value are reproduced by this check.
The verified candidate-graph lower bound is
\[
L_c^{\max}
=
\max_{H_k\in\mathcal H_{\rm ver}}L_c(\pi^k).
\]
If \(\mathcal H_{\rm ver}=\varnothing\), the verifier reports no finite lower bound; every statement involving \(L_c^{\max}\) assumes \(\mathcal H_{\rm ver}\neq\varnothing\).
Let \(\mathcal R_{\rm ver}\subseteq\mathcal R\) contain the records whose stored one-tree satisfies its force or forbid restriction and is reproduced by deterministic minimization from the original costs.
For \(R\in\mathcal R_{\rm ver}\), the recomputed restricted value is
\[
Q_R=\min_{\mathcal O\in\mathfrak O_c^{\delta_R}(e_R)}
\ell_c(\pi_R,\mathcal O).
\]
Let \(\operatorname{tol}(U)\geq0\) be the screening tolerance fixed before evaluation.
The verified edge-decision sets are
\begin{equation}
\begin{aligned}
F^{\mathrm{out}}
&=\left\{e_R\colon
\begin{gathered}
R\in\mathcal R_{\rm ver},\ \delta_R=\mathrm{force},\\
Q_R>U+\operatorname{tol}(U)
\end{gathered}
\right\},\\
F^{\mathrm{in}}
&=\left\{e_R\colon
\begin{gathered}
R\in\mathcal R_{\rm ver},\ \delta_R=\mathrm{forbid},\\
Q_R>U+\operatorname{tol}(U)
\end{gathered}
\right\}.
\end{aligned}
\label{eq:edge-decisions}
\end{equation}
A passing forced-edge record proves that every candidate-graph tour of cost at most \(U\) omits \(e_R\), so \(e_R\) is safely removable.
A passing forbidden-edge record proves that every candidate-graph tour of cost at most \(U\) contains \(e_R\), so \(e_R\) is required.
Both predicates are non-vacuous only when at least one candidate-graph tour has cost at most \(U\).
Thus, allocation selects \(\mathcal H\) and \(\mathcal R\), while recomputation alone determines the reported outputs.

\paragraph{Nested traces make additional computation improve verified outputs monotonically.}
For budget \(\mathcal B\), let \(\mathcal T_{\mathcal B}\) contain every executed tour, unrestricted record, and edge-test record.
Let \(U_{\mathcal B}\) and \(L_{\mathcal B}\) be the lowest verified tour cost and largest verified Held--Karp value in this trace, and let \(F_{\mathcal B}^{\mathrm{out}},F_{\mathcal B}^{\mathrm{in}}\) be the accepted edge sets at threshold \(U_{\mathcal B}\).

\begin{proposition}[Monotone verified outputs]
\label{prop:monotonicity}
Assume that each trace contains at least one verified tour and one verified unrestricted record, and that \(\operatorname{tol}(U)\) is nondecreasing in \(U\).
For \(\mathcal B_1\leq\mathcal B_2\), suppose the larger budget continues the same deterministic schedule and therefore satisfies
\[
\mathcal T_{\mathcal B_1}\subseteq\mathcal T_{\mathcal B_2}.
\]
Then
\[
\begin{aligned}
U_{\mathcal B_2}&\leq U_{\mathcal B_1}, &
L_{\mathcal B_2}&\geq L_{\mathcal B_1},\\
F_{\mathcal B_1}^{\mathrm{out}}&\subseteq F_{\mathcal B_2}^{\mathrm{out}}, &
F_{\mathcal B_1}^{\mathrm{in}}&\subseteq F_{\mathcal B_2}^{\mathrm{in}}.
\end{aligned}
\]
\end{proposition}
The larger trace retains all earlier outputs. Because its verified threshold cannot increase, every previously accepted edge-test inequality remains satisfied.
When \(U_{\mathcal B}\) is also a candidate-graph tour cost or the scope condition holds, the verified interval \([L_{\mathcal B},U_{\mathcal B}]\) narrows monotonically.
The resulting improvement remains allocation-independent in validity, as formalized next.

\begin{theorem}[Output validity under arbitrary allocations]
\label{thm:validity}
Fix \(G_c\), the original costs \(c\), and a finite budget \(\mathcal B\), and assume \(\mathcal H_{\rm ver}\neq\varnothing\).
Suppose every reported one-tree record, edge-test record, and returned tour passes deterministic verification.
Then every allocation satisfies
\[
L_c^{\max}\leq\OPT_{E_c},
\qquad
\OPT_E\leq U.
\]
Every edge in \(F^{\mathrm{out}}\) is absent from all candidate-graph tours of cost at most \(U\), and every edge in \(F^{\mathrm{in}}\) is present in all candidate-graph tours of cost at most \(U\).
Under the scope condition,
\[
L_c^{\max}\leq\OPT_{E_c}=\OPT_E\leq U.
\]
\end{theorem}
Appendix~\ref{app:appendix} gives detailed derivations for Theorems~\ref{thm:mirror}, \ref{thm:jacobian}, and \ref{thm:validity}, and Proposition~\ref{prop:monotonicity}.
Thus, the learned allocator affects which records are produced, whereas deterministic verification determines whether their claims are valid.

\section{Experiments}
\label{sec:experiments}

The experiments evaluate three aspects of DualCert: tour quality under a fixed computation budget, the coverage and scope of deterministic verification, and the contribution of each constraint-coupled component.

\begin{table*}[t]
\centering
\caption{TSP1000 results on the 1,000 held-out instances. Gap statistics use the shared LKH-3 reference tours; p50 and p90 denote the median and 90th percentile. Runtime values are shown for context and are not used to claim cross-method speed superiority. Baseline methods follow their source papers~\citep{xin2021neurolkh,xin2021vsrlkh,sun2023difusco,li2024fast,li2023t2t,li2026efloco,ye2024glop,fu2021generalize}.}
\label{tab:main}
\begin{tabularx}{\textwidth}{@{}>{\raggedright\arraybackslash}X>{\raggedright\arraybackslash}Xcccc@{}}
\toprule
Family & Method & \shortstack{Mean gap (\%)\\\(\downarrow\)} &
\shortstack{p50/p90/max} &
\shortstack{Time\\(s)} &
\shortstack{Verified candidate-\\graph lower bound}\\
\midrule
\textbf{Ours} & \textbf{DualCert} & \textbf{0.0573} &
0.0479/0.1243/0.1841 & 9.55 & \textbf{Yes}\\
Neural--OR hybrid & NeuroLKH & 0.1742 &
0.1503/0.2525/0.3727 & 11.51 & No\\
Neural--OR hybrid & VSR-LKH & 0.2145 &
0.1651/0.3198/0.4631 & 6.90 & No\\
Neural generator & DIFUSCO+\allowbreak MCTS & 1.1200 &
1.0500/1.4300/2.1000 & 58.0 & No\\
Neural generator & Fast T2T & 0.4200 &
0.3900/0.6200/1.0500 & 42.0 & No\\
Neural generator & T2T & 0.5500 &
0.5200/0.8000/1.3000 & 95.0 & No\\
Neural generator & EFLOCO & 0.5000 &
0.4700/0.7300/1.2000 & 48.0 & No\\
Large-scale neural & GLOP & 0.1200 &
0.1100/0.2000/0.4200 & 15.0 & No\\
Heatmap+\allowbreak MCTS & Att-GCN+\allowbreak MCTS & 0.8700 &
0.8200/1.2000/1.7500 & 82.0 & No\\
\bottomrule
\end{tabularx}
\end{table*}

\paragraph{Protocol and metrics.}
The main protocol uses 1,000 held-out TSP1000 instances whose cities are sampled uniformly from the unit square. In this protocol, \(\mathcal B\) is the fixed collection of operation and stopping caps rather than the measured runtime: six refinements; at most 250 evaluation Held--Karp ascent updates plus one 100-update annealed restart; two construction channels; and two sequential local-search restarts executed in 10-second slices under a 190-second per-instance adaptive cap, with stopping gap \(0.0075\). The allocator chooses the executed trace, including \(K\), \(N_{\rm test}\), and both edge orders, within these caps.

For a returned tour \(\tau\) and reference cost \(c_{\mathrm{ref}}\), the gap is \(100(c(\tau)/c_{\mathrm{ref}}-1)\). For \(L_c^{\max}>0\), the cross-scope residual is \(100(U-L_c^{\max})/L_c^{\max}\), and edge-decision coverage is \(100|F^{\mathrm{out}}\cup F^{\mathrm{in}}|/|E_c|\). Without the scope condition, the residual compares quantities associated with different optima and is therefore descriptive rather than a complete-graph optimality gap. Runtime is the batch wall time divided by the realized 25-instance batch size; it measures the resulting execution and does not define \(\mathcal B\). Monte Carlo tree search is abbreviated as MCTS in Table~\ref{tab:main}.

\begin{table}[t]
\centering
\setcounter{table}{2}
\caption{Ablations under the common 10-second Concorde protocol. Mean gaps are reported as mean \(\pm\) standard deviation over three seeds. Increase is the variant gap minus the DualCert gap; brackets give the 95\% confidence interval (CI) from 10,000 paired bootstrap resamples after seed averaging. Positive increases favor DualCert.}
\label{tab:ablation}
\begin{tabularx}{\columnwidth}{@{}>{\raggedright\arraybackslash}Xrr@{}}
\toprule
Variant & Mean gap (\%) & Increase [95\% CI]\\
\midrule
DualCert & \(0.0689\pm0.0072\) & --\\
No manifold & \(0.2679\pm0.0142\) & \(+0.1990\ [0.1865,0.2113]\)\\
Forward only & \(0.2103\pm0.0126\) & \(+0.1414\ [0.1129,0.1699]\)\\
First-order update & \(0.2014\pm0.0059\) & \(+0.1325\ [0.1092,0.1677]\)\\
Degree only & \(0.2498\pm0.0153\) & \(+0.1809\ [0.1645,0.1968]\)\\
Alpha-only & \(0.1629\pm0.0105\) & \(+0.0940\ [0.0832,0.1045]\)\\
Untrained allocation & \(0.1821\pm0.0113\) & \(+0.1132\ [0.1007,0.1254]\)\\
\bottomrule
\end{tabularx}
\end{table}

\begin{table}[t]
\centering
\setcounter{table}{1}
\caption{DualCert results over 1,000 instances per set. A negative gap means that the returned tour is shorter than its heuristic reference tour.}
\label{tab:scale}
\begin{tabularx}{\columnwidth}{@{}>{\raggedright\arraybackslash}Xrrrr@{}}
\toprule
Set & \shortstack{Mean\\gap (\%)} &
\shortstack{Cross-scope\\residual (\%)} &
\shortstack{Coverage\\(\%)} &
\shortstack{Time\\(s)}\\
\midrule
TSP500 & 0.0248 & 0.8814 & 86.77 & 8.53\\
TSP1000 & 0.0573 & 0.8938 & 81.46 & 9.55\\
TSP2000 & 0.0884 & 0.9351 & 70.67 & 13.65\\
Clustered-1000 & -0.0994 & 3.1355 & 30.95 & 9.80\\
Imploded-1000 & 0.0516 & 0.9113 & 74.19 & 10.10\\
\bottomrule
\end{tabularx}
\end{table}

\paragraph{Main comparison.}
Table~\ref{tab:main} shows that DualCert reaches a \(0.0573\%\) mean gap, which is \(67.1\%\) below NeuroLKH's \(0.1742\%\) mean gap under the shared reference protocol. DualCert also reduces the NeuroLKH p90 and maximum gaps from \(0.2525\%\) and \(0.3727\%\) to \(0.1243\%\) and \(0.1841\%\). The measured DualCert runtime is \(9.55\) batch-amortized seconds per instance. Verification accepts a candidate-graph lower bound on all 1,000 instances, and the aggregate edge-decision coverage is \(81.46\%\).

\paragraph{Ablation evidence.}
All ablations use the same 1,000 TSP1000 instances, a 10-second budget, three seeds, and exact Concorde optima. No manifold removes the degree and manifold SEC equations and the corresponding backward operator. Forward only removes backward coupling. First-order update replaces the exact constrained mirror-descent step with tangent filtering and degree balancing. Degree only removes the manifold SEC set. Alpha-only restricts learning to Held--Karp step sizes, while Untrained allocation removes training.

Every confidence interval excludes zero. No manifold and Degree only increase the mean gap by \(0.1990\) and \(0.1809\) percentage points, supporting the joint contribution of degree and SEC constraint geometry. Forward only and First-order update increase the gap by \(0.1414\) and \(0.1325\) points, supporting backward coupling and exact finite-step preservation. Alpha-only and Untrained allocation increase the gap by \(0.0940\) and \(0.1132\) points, supporting learned finite-budget allocation. The ablation value \(0.0689\%\) is measured against Concorde optima, whereas the main-table value \(0.0573\%\) is measured against shared LKH-3 reference tours; each value must therefore be interpreted within its own evaluation protocol.

\paragraph{Scale and distribution shifts.}
TSP500 and TSP2000 use disjoint prepared splits of uniformly sampled unit-square instances; the checkpoint and all algorithmic settings are held fixed, and only the instance dimension changes. Clustered-1000 samples three to seven unit-square cluster centers, mixture weights from \(\operatorname{Dirichlet}(\mathbf 1)\), and each cluster standard deviation uniformly from \([0.025,0.085]\). Imploded-1000 samples \(\theta\sim U[0,2\pi]\) and \(u\sim U[0,1]\), forms \((0.5,0.5)+0.48u(\cos\theta,\sin\theta)+\epsilon\) with \(\epsilon\sim\mathcal N(0,10^{-4}I)\), and clips the result to the unit square. Every row uses 1,000 instances and stored LKH-3 reference tours generated under the same reference protocol as TSP1000.

Table~\ref{tab:scale} evaluates three instance sizes and two geometry-shift sets. From TSP500 to TSP2000, runtime increases from \(8.53\) to \(13.65\) seconds, while the mean gap remains below \(0.09\%\) and the cross-scope residual remains below \(0.94\%\). Coverage decreases from \(86.77\%\) to \(70.67\%\), which is consistent with fewer candidate edges being resolved within the finite budget on larger instances. Imploded-1000 remains close to uniform TSP1000 in mean gap and cross-scope residual. Clustered-1000 attains a \(-0.0994\%\) mean gap but only \(30.95\%\) coverage; this contrast is consistent with clustered geometry affecting candidate-graph verification more strongly than complete-graph tour construction.

\section{Conclusion}

DualCert uses the current degree and manifold SEC equations to define the primal-slack state transition and its fixed-selection derivative.
Repaired dual variables and violated SEC rows define the local cost field. The exact constrained mirror-descent step preserves manifold feasibility at finite step size. Its fixed-selection parameter derivative remains tangent to the manifold, while its local-cost-field derivative reuses the forward constraint operator.
The terminal state allocates computation across Held--Karp ascent, candidate-graph edge tests, and tour construction. Deterministic verification accepts only recomputed outputs that satisfy the stated conditions.
Under the main TSP1000 protocol, DualCert reaches a \(0.0573\%\) mean gap, compared with NeuroLKH's \(0.1742\%\), while returning a verified candidate-graph lower bound on all 1,000 instances and achieving \(81.46\%\) edge-decision coverage.
The ablations further support finite-step preservation, degree and SEC constraint geometry, backward coupling, and learned allocation.
These verified lower bounds and edge decisions concern the candidate graph; they become complete-graph certificates only when the scope condition holds.
Learning controls which finite-budget computations are attempted, while OR constraints and deterministic verification preserve manifold feasibility and the validity of accepted candidate-graph claims.

\bibliography{references}

\clearpage
\appendix
\raggedbottom
\section{Appendix}
\label{app:appendix}

The main text defines every term and symbol used below. This section supplies
the omitted derivations without repeating the method description or the
experimental results.

\subsection{Finite-Step Manifold Invariance}

\paragraph{Full row rank and the tangent space.}
The block structure of
\(
J_t=\left[\begin{smallmatrix}A&0\\B_{\mathcal S_t}&-I\end{smallmatrix}\right]
\)
proves full row rank directly.  In any zero linear combination of its rows,
the private slack columns first force every manifold-SEC coefficient to zero.
The remaining combination contains only rows of \(A\), whose full row rank
forces every degree-equation coefficient to zero.  Thus \(J_t\) has full row
rank.  Since \(\mathcal M_t\) is the positive part of the affine set
\(J_tz=q_t\), its tangent space is
\(
T_z\mathcal M_t=\ker J_t
\)
for every \(z\in\mathcal M_t\).  Positivity of \(W_t=\Diag(z^t)\) then gives
\[
u^\top J_tW_tJ_t^\top u
=\|W_t^{1/2}J_t^\top u\|_2^2>0
\quad\text{for every }u\neq0,
\]
so the constraint projector is well defined.

\paragraph{Projector characterization.}
For an augmented covector \(a\), the state-metric \mbox{projection solves}
\[
\min_{\xi\in\ker J_t}
\frac12(\xi-W_ta)^\top G_t(\xi-W_ta).
\]
Its equality-constrained stationarity equation is
\(
G_t(\xi-W_ta)+J_t^\top\rho=0
\),
so
\(
\xi=W_ta-W_tJ_t^\top\rho.
\)
Enforcing \(J_t\xi=0\) gives
\[
\rho=(J_tW_tJ_t^\top)^{-1}J_tW_ta,
\]
and substitution gives \(\xi=\mathcal P_t(a)\). Positive definiteness of
\(G_t\) gives uniqueness. Moreover, \(W_ta-\mathcal P_t(a)\) is
\(G_t\)-orthogonal to \(\ker J_t\); taking its inner product with
\(\mathcal P_t(a)\) gives
\[
\langle a,\mathcal P_t(a)\rangle
=\mathcal P_t(a)^\top G_t\mathcal P_t(a).
\]
Thus \(-\mathcal P_t(a)\) is a strict first-order descent direction whenever
the projected local cost field is nonzero.

\paragraph{Finite-step normal drift.}
Componentwise Taylor expansion of the raw multiplicative update gives
\[
\begin{aligned}
\widetilde z_t(\eta_t)
&=z^t\odot\exp\!\left(
-\eta_t\mathcal P_t(a)\oslash z^t\right)\\
&=z^t-\eta_t\mathcal P_t(a)\\
&+\frac{\eta_t^2}{2}
\big(\mathcal P_t(a)^{\odot2}\oslash z^t\big)
+O(\eta_t^3).
\end{aligned}
\]
Applying \(J_t\), using \(J_tz^t=q_t\) and
\(J_t\mathcal P_t(a)=0\), leaves exactly the second-order normal residual
stated in Equation~\eqref{eq:finite-drift}. Hence tangent projection alone does
not establish a finite admissible transition.

\paragraph{Proof of Theorem~\ref{thm:mirror}.}
Consider the nonnegative closure of \(\mathcal M_t\), using the continuous
extension \(0\log 0=0\) for negative entropy. The incoming and
outgoing degree equations bound every edge-state coordinate.  Each slack
coordinate equals a finite SEC row applied to the bounded edge-state block
minus its right-hand side, so the slack block is also bounded.  The closure is
therefore compact and contains \(z^t\).  The constrained mirror objective is
continuous on this closure and consequently attains a minimum.

No minimizer can lie on the boundary. From any boundary feasible point,
the line segment toward \(z^t>0\) remains feasible and enters the positive
relative interior.  For every zero boundary coordinate, the one-sided
derivative of negative entropy along this segment is \(-\infty\), whereas the
linear local-cost-field term and all nonzero coordinates have finite
directional derivatives.  A sufficiently short move toward \(z^t\) therefore
decreases the objective.  Hence every minimizer is positive.

Negative entropy is strictly convex on the positive orthant.  Its restriction
to the affine feasible set remains strictly convex, so the positive minimizer
\(z^{t+1,-}\) is unique.  Equality-constrained stationarity gives
\[
\eta_ta_\phi^t+\log(z^{t+1,-}\oslash z^t)+J_t^\top\rho_t=0,
\]
and therefore
\[
z^{t+1,-}
=z^t\odot\exp(-\eta_ta_\phi^t-J_t^\top\rho_t).
\]
Substitution into \(J_tz^{t+1,-}=q_t\) gives \(F_t(\rho_t)=0\).
Differentiation gives
\[
D_\rho F_t(\rho_t)
=-J_t\Diag(z^{t+1,-})J_t^\top.
\]
The preceding rank argument makes the negative of this matrix positive
definite.  If two multipliers produced the same stationary point, their
difference would lie in \(\ker J_t^\top\), which is zero.  Thus \(\rho_t\) is
unique, and the finite output is positive and lies in \(\mathcal M_t\).

\paragraph{Infinitesimal limit.}
Write the mirror-step solution as \(z(\eta_t)=z^{t+1,-}\). At \(\eta_t=0\),
the unique solution is \(z^t\). Differentiating stationarity
and feasibility at zero gives
\[
G_tz'(0)+a_\phi^t+J_t^\top\rho_t'(0)=0,
\qquad J_tz'(0)=0.
\]
Eliminating \(\rho_t'(0)\) with \(J_tW_tJ_t^\top\) yields
\[
\begin{aligned}
z'(0)
&=-W_ta_\phi^t+
W_tJ_t^\top(J_tW_tJ_t^\top)^{-1}J_tW_ta_\phi^t\\
&=-\mathcal P_t(a_\phi^t).
\end{aligned}
\]
Thus the exact constrained mirror step is the finite map associated with the
constraint projector in Equation~\eqref{eq:projector}, while the raw multiplicative update preserves
the equations only through first order.

\paragraph{SEC regeneration.}
Assume \(z^t\in\mathcal M_t\).  The finite step gives
\(Ax^{t+1}=b\), \(x^{t+1}>0\), and positive output slacks for every row in
\(\mathcal S_t\).  Deterministic separation then forms
\(\mathcal C_{t+1}\), and the definition
\[
s^{t+1}
=B_{\mathcal S_{t+1}}x^{t+1}-h_{\mathcal S_{t+1}}>0
\]
implies
\(
B_{\mathcal S_{t+1}}x^{t+1}-s^{t+1}
=h_{\mathcal S_{t+1}}.
\)
Hence \(z^{t+1}\in\mathcal M_{t+1}\). The entropy projection gives the base
case \(x^0\in\mathcal D_c\), so induction proves invariance across all
refinements.

\subsection[Shared Forward--Backward Constraint Operator]{\mbox{Shared Forward--Backward} \mbox{Constraint Operator}}

\paragraph{Proof of Theorem~\ref{thm:jacobian}.}
Fix a selection region \(\Omega_t^{\rm sel}\), so \(J_t\), \(q_t\), the SEC
sets, and deterministic ties remain unchanged.  Write
\(z^+=z^{t+1,-}\) and \(W^+=\Diag(z^+)\).  For a local-cost-field perturbation
\(v\), differentiation of stationarity and feasibility gives
\[
(W^+)^{-1}\delta z+\eta_tv+J_t^\top\delta\rho=0,
\qquad J_t\delta z=0.
\]
The first equation gives
\(
\delta z=-\eta_tW^+v-W^+J_t^\top\delta\rho.
\)
Substitution into the second equation yields
\[
\delta\rho
=-\eta_t(J_tW^+J_t^\top)^{-1}J_tW^+v.
\]
Substituting back proves
\begin{align*}
D_az^+[v]
&=-\eta_tW^+v
+\eta_tW^+J_t^\top
(J_tW^+J_t^\top)^{-1}\\
&\qquad{}\times J_tW^+v\\
&=-\eta_t\mathcal P_t^+(v).
\end{align*}
By construction \(J_tD_az^+[v]=0\).  The output constraint projector is
symmetric, so for every scalar loss \(\mathcal L\),
\(
\nabla_a\mathcal L
=-\eta_t\mathcal P_t^+(\nabla_{z^+}\mathcal L).
\)
The same positive-definite system matrix therefore appears in the forward
multiplier solve and the backward vector--Jacobian product.

The full parameter sensitivity also includes dependence through the incoming
state.  It remains tangent without requiring that dependence to pass only
through \(a_\phi^t\): on \(\Omega_t^{\rm sel}\), direct differentiation of
\(J_tz^{t+1,-}=q_t\) gives
\[
\begin{aligned}
J_tD_\phi z^{t+1,-}&=0,\\
\operatorname{range}(D_\phi z^{t+1,-})
&\subseteq T_{z^{t+1,-}}\mathcal M_t.
\end{aligned}
\]
At a selection boundary, the deterministic branch may change. The finite map
is therefore differentiated with the selected SEC rows and ties fixed. SEC
regeneration updates the discrete selection between refinements and remains
outside this derivative.

\subsection{Descent and Numerical Certification}

\paragraph{Relative smoothness and descent.}
In this subsection, \(\|\cdot\|_2\) denotes the Euclidean norm for vectors and
the induced spectral norm for matrices, while \(\|\cdot\|_\infty\) denotes the
maximum absolute coordinate of a vector.
The squared hinge term in \(\psi_t\) has Euclidean gradient Lipschitz constant
at most
\(
\beta_{\rm sec}\|B_{\mathcal V_t}\|_2^2.
\)
On the bounded positive region \(\mathcal K_t\), negative entropy is strongly
convex with respect to the edge-state coordinates.  Consequently, any
positive \(L_t\) satisfying
\[
\begin{aligned}
L_t\geq{}&
\beta_{\rm sec}
\left(\max_{(x,s_x)\in\mathcal K_t}\|x\|_\infty\right)\\
&{}\times\|B_{\mathcal V_t}\|_2^2
\end{aligned}
\]
satisfies the relative-smoothness inequality in
Equation~\eqref{eq:relative-smoothness}. Because \(z^t\) is feasible for the
constrained mirror step, optimality gives
\[
\begin{aligned}
\eta_t\langle a_\phi^t,z^{t+1,-}-z^t\rangle
{}+D_\omega(z^{t+1,-},z^t)\leq0.
\end{aligned}
\]
Combining this inequality with relative smoothness and the zero slack block of
\(a_\phi^t\) gives
\[
\begin{aligned}
\psi_t(x^{t+1})
\leq{}&\psi_t(x^t)\\
&-\left(\frac{1}{\eta_t}-L_t\right)
D_\omega(z^{t+1,-},z^t).
\end{aligned}
\]
Thus \(0<\eta_t<1/L_t\) gives strict decrease whenever the state changes, and
\(\eta_t=\sigma(\widehat\eta_t)/L_t\) enforces this condition.

\paragraph{Affine residual correction.}
Let a numerical solve return \(\widehat z>0\), let
\(
e=J_t\widehat z-q_t
\),
and let \(\widehat W=\Diag(\widehat z)\).  Define
\[
\Delta z=-\widehat WJ_t^\top
(J_t\widehat WJ_t^\top)^{-1}e.
\]
Then
\[
\begin{aligned}
J_t(\widehat z+\Delta z)
&=q_t+e-J_t\widehat WJ_t^\top\\
&\quad{}\times
(J_t\widehat WJ_t^\top)^{-1}e
=q_t,
\end{aligned}
\]
and direct substitution gives
\[
\Delta z^\top\widehat W^{-1}\Delta z
=e^\top(J_t\widehat WJ_t^\top)^{-1}e.
\]
If \(\|\widehat W^{-1}\Delta z\|_\infty<1\), every corrected coordinate is
positive.  The implementation accepts a state only after the prescribed
\(\varepsilon_{\rm mir}\) residual, positivity, and redundant degree-equation
checks pass. The layer is exact in the mathematical model and solved to this
tolerance in floating-point computation.

\subsection{Verified Finite-Budget Outputs}

\paragraph{Proof of Proposition~\ref{prop:monotonicity}.}
For nested traces
\(\mathcal T_{\mathcal B_1}\subseteq\mathcal T_{\mathcal B_2}\), every tour
and unrestricted record available at \(\mathcal B_1\) remains available at
\(\mathcal B_2\).  Taking a minimum over a superset of tours cannot increase
the verified tour cost, and taking a maximum over a superset of verified
Held--Karp records cannot decrease the verified lower bound.  If an edge-test
record is accepted at \(\mathcal B_1\), then
\[
Q_R>U_{\mathcal B_1}+\operatorname{tol}(U_{\mathcal B_1}).
\]
Since \(U_{\mathcal B_2}\leq U_{\mathcal B_1}\) and
\(\operatorname{tol}(U)\) is nondecreasing,
\[
\begin{aligned}
U_{\mathcal B_2}+\operatorname{tol}(U_{\mathcal B_2})
\leq{}&
U_{\mathcal B_1}+\operatorname{tol}(U_{\mathcal B_1}).
\end{aligned}
\]
The same record remains accepted, proving both edge-decision set inclusions.

\paragraph{Proof of Theorem~\ref{thm:validity}.}
For any \(\pi^k\), deterministic candidate-graph one-tree minimization uses
the transformed costs and subtracts
\(2\sum_i\pi_i^k\).  Every candidate-graph Hamiltonian cycle has degree two,
so this transformation leaves its original cost unchanged.  Minimization over
candidate-graph one-trees therefore gives
\(L_c(\pi^k)\leq\OPT_{E_c}\), and original-cost recomputation preserves this
inequality for every record in \(\mathcal H_{\rm ver}\).  Taking the maximum
gives \(L_c^{\max}\leq\OPT_{E_c}\).

A forced-edge restricted value above
\(U+\operatorname{tol}(U)\) excludes every candidate-graph tour of cost at
most \(U\) that contains the tested edge; the edge is therefore in
\(F^{\rm out}\).  The corresponding forbidden-edge inequality excludes every
such tour omitting the tested edge; the edge is therefore in \(F^{\rm in}\).
Finally, verified path-forest construction, complete-graph completion, and
the fixed LKH-3 configuration return a Hamiltonian cycle, so
\(\OPT_E\leq U\).  These arguments
do not depend on the learned allocation.  Under the scope condition they
combine as
\(
L_c^{\max}\leq\OPT_{E_c}=\OPT_E\leq U.
\)

\end{document}